\documentclass[lettersize,journal]{IEEEtran}
\usepackage{amsmath,amsfonts}
\usepackage{algorithmic}
\usepackage{algorithm}
\usepackage{array}
\usepackage[caption=false,font=normalsize,labelfont=sf,textfont=sf]{subfig}
\usepackage{textcomp}
\usepackage{stfloats}
\usepackage{url}
\usepackage{verbatim}
\usepackage{graphicx}
\usepackage{cite}
\begin{document}

\title{Out-of-distribution Generalization via Partial Feature Decorrelation}

\author{Xin Guo*, Zhengxu Yu*, Chao Xiang, Zhongming Jin, Jianqiang Huang, Deng Cai, \IEEEmembership{Senior Member,~IEEE,} Xiaofei He, \IEEEmembership{Senior Member,~IEEE,} and Xian-Sheng Hua, \IEEEmembership{Fellow, IEEE}% <-this % stops a space

\thanks{Xin Guo, Chao Xiang, Deng Cai, and Xiaofei He are with the State Key Lab of CAD\&CG, Zhejiang University, Hangzhou 310058, P.R.China  e-mail: guoxinzju@gmail.com, chaoxiang@zju.edu.cn, \{dengcai, xiaofeihe\}@cad.zju.edu.cn}

\thanks{Zhengxu Yu, Zhongming Jin, Jianqiang Huang, and Xian-Sheng Hua are with the DAMO Academy, Alibaba Group, Hangzhou 311121, P.R.China e-mail: \{yuzxfred, huaxiansheng\}@gmail.com \{zhongming.jinzm, jianqiang.hjq\}@alibaba-inc.com}

\thanks{'*' means equal contribution.}
}

\markboth{Journal of \LaTeX\ Class Files,~Vol.~14, No.~8, August~2021}%
{Shell \MakeLowercase{\textit{et al.}}: A Sample Article Using IEEEtran.cls for IEEE Journals}


\maketitle

\begin{abstract}
    Most deep-learning-based image classification methods assume that all samples are generated under an independent and identically distributed (IID) setting. However, out-of-distribution (OOD) generalization is more common in practice, which means an agnostic context distribution shift between training and testing environments. To address this problem, we present a novel Partial Feature Decorrelation Learning (PFDL) algorithm, which jointly optimizes a feature decomposition network and the target image classification model. The feature decomposition network decomposes feature embeddings into the independent and the correlated parts such that the correlations between features will be highlighted. Then, the correlated features help learn a stable feature representation by decorrelating the highlighted correlations while optimizing the image classification model. We verify the correlation modeling ability of the feature decomposition network on a synthetic dataset. The experiments on real-world datasets demonstrate that our method can improve the backbone model's accuracy on OOD image classification datasets.
\end{abstract}


\section{Introduction}
    \IEEEPARstart{C}{lassifying} unknown images based on a model trained on a training dataset is a common machine learning problem~\cite{lecun1998gradient,krizhevsky2012imagenet,szegedy2015going,simonyan2014very,he2016deep}. With the rise of deep learning methods, image classification accuracy has been improved tremendously on many public datasets~\cite{deng2009imagenet,krizhevsky2009learning,deng2012mnist,fei2004learning,everingham2015pascal}. Most deep-learning-based image classification methods follow a consistency assumption that all samples are generated under an independent and identically distributed (IID) setting~\cite{he2021towards}. Under this assumption, an approximation of the true distribution can be learned by yielding a lower loss on the training set. However, in real-world applications, we cannot guarantee the unknown test data have the same distribution as the training data due to the data generation bias \cite{hua2021feature,liu2021gafnet,guo2017robust, guo2021discriminative,liu2020feature,he2021towards}, which is known as out-of-distribution (OOD) generalization\cite{nagarajan2020understanding,xie2020n,bai2020decaug}. Consequently, in such training environment, these IID-based methods will inevitably overfit the statistical correlations between features and omit the actual causality between the target object and the label \cite{torralba2011unbiased}.
    
    \begin{figure}[t]
        \centering
        \includegraphics[width=0.98\linewidth]{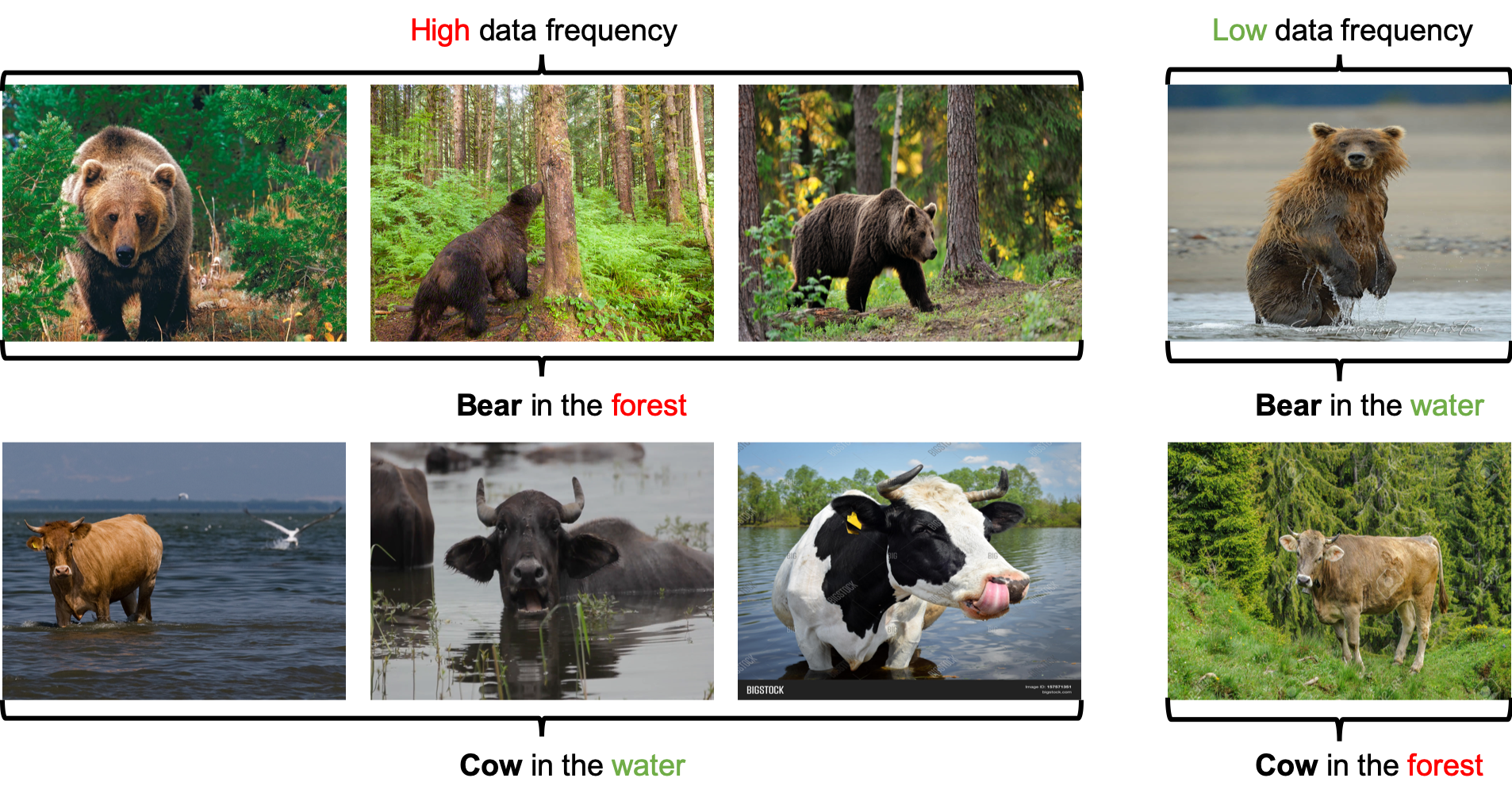}
        \caption{Illustration of context distribution shifting between training and testing datasets. Image source: NICO dataset \protect \cite{he2021towards}. }
        \label{fig:case1}
    \end{figure}
    
    A toy example of this agnostic distribution shift problem caused by data generation bias has been shown in Figure.~\ref{fig:case1}, in which 'bear in the forest' has a higher data frequency than 'bear in the water.' The label of the first row is 'bear', and the second row is 'cow'. A typical IID-based deep-learning model trained on such a dataset will inevitably absorb the correlation between object 'bear' and context 'forest' to yield a lower training loss. Consequently, causality between the feature of the target object 'bear' and the label 'bear' is blurred by the correlation between the feature of context 'forest' and the label 'bear'. We can find a similar correlation between the context feature 'water' and the label 'cow' in Figure.~\ref{fig:case1}. This dilemma can lead to an unstable performance in a testing environment where 'cow in the forest' has higher data frequency. Because the misspecified model partly relies on the correlation between the context features 'water' and the label 'cow'.
    
    Recently, many works have been proposed to solve this agnostic distribution shift problem \cite{wang2020decorrelated,pan2018macnet,wang2019minimax,shen2020stable,kuang2020stable,zhang2021deep}, including domain adaption-based methods and some causality-based sample re-weighting methods \cite{chen2020adversarial,pan2018macnet,wang2019minimax}. Some domain adaption-based methods are based on a straightforward thought of taking advantage of the prior knowledge of the testing environment. However, it is impossible to obtain prior knowledge of the test data in many real-world applications, which hinders the application of these methods. As for causality-based sample re-weighting methods, most of them \cite{shen2018causally,shen2020stable,kuang2020stable,arjovsky2020invariant} are based on increasing the importance of samples with lower data frequency to mitigate the impact of the agnostic distribution shift problem. Most of these methods require a large re-weighting matrix whose parameter number is proportional to the number of training samples, both computation and memory intensive.
    
    To address these problems, we propose to learn a stable feature representation in an OOD generalization setting by decomposing and removing the effect of correlations on the feature embeddings. Specifically, we propose a novel Partial Feature Decorrelation Learning (PFDL) algorithm for OOD image classification tasks. In this algorithm, we jointly optimize a feature decomposition network and the target image classification model while fixing each of the network's parameters in turn. The start point of the feature decomposition network is to learn and decompose the correlated part features from the learned feature embedding using the followed decorrelation operations. After that, we decorrelate the correlated part features while optimizing the target image classification model. We can learn a stable feature representation by decorrelating as many correlations as possible without disturbing the stable connections between features. Compared with previous sample re-weighting-based methods, our model's parameter number is not proportional to the training sample number, which provides better scalability than previous works in applications with massive training data. The experimental results demonstrate that the model's accuracy and stability trained by our method can outperform all the baselines and the state-of-the-art stable learning methods on several OOD datasets.
    
    We summarize the contributions of this work as follows:
    
    (1) We study how to learn a stable feature representation under OOD generalization problems from decomposing and decorrelating the correlated part features to improve the model's accuracy and propose a novel Partial Feature Decorrelation Learning (PFDL) algorithm.
    
    (2) We design a feature decomposition network that separates the independent features from the correlated ones. We also develop a Partial Feature Decorrelation objective function, which aims to learn a stable feature representation by decorrelating the correlations while optimizing the target image classification model.
    
    (3) We verify the correlation modeling ability of the feature decomposition network on a synthetic dataset. The experimental results demonstrate that our model can achieve state-of-the-art performance on real-world datasets.

\section{Related Work}
\subsection{Causality-based Methods}
    Most recently proposed causality-based methods \cite{arjovsky2020invariant,shen2018causally,shen2020stable,kuang2020stable,peters2016causal} are based on sample re-weighting, which does not directly change the biased sample features but shifts the training dataset's distribution by varying the importance of the samples. \cite{shen2020stable} proposes a sample re-weighting method to address the collinearity among input variables caused by the agnostic distribution shift. \cite{kuang2020stable} proposes a feature decorrelation-based sample re-weighting method to decorrelate all features without considering the possible causality between features. However, these sample re-weighting methods require a large re-weighting matrix whose parameter number is proportional to the training samples. Hence, these works are both computation and memory intensive. This disadvantage limits their scalability in machine learning tasks with a large number of training data. Unlike these methods, our model learns the associations between features using a neural network whose parameter number is not proportional to the sample number, which provides feasibility for tasks with an extensive training dataset.
    
\subsection{Non-causality-based Methods}
    In addition to causality-based methods, a variety of domain adaptation \cite{chen2020adversarial,liu2014robust,zadrozny2004learning}, domain generalization \cite{muandet2013domain} and transfer learning methods \cite{pan2018macnet,wang2019minimax} are proposed to address the OOD generalization problem. Some of these methods handle the distribution shift between training and testing datasets by aligning the training dataset to the target dataset or vice versa. To achieve that, these methods require prior knowledge of the target domain distribution. However, it is impossible to acquire prior knowledge in many real-world applications. Unlike them, our method is intended to partition the observed associations among features to improve the model's performance, which requires no prior knowledge of the testing environments.

\subsection{Feature Decorrelation}
    Some Lasso-based framework~\cite{chen2013uncorrelated,takada2018independently} propose to decorrelate features by adding a regularizer that imposes the highly correlated features not to be selected simultaneously. \cite{kuang2020stable,shen2020stable,zhang2021deep} propose to address the problem by removing the dependencies between features via learning weights for training samples, which helps deep models get rid of spurious correlations and, in turn, concentrate more on the true connection between discriminative features and labels.

\section{Partial Feature Decorrelation Learning}
    Our goal is to learn a predictive model for image classification with agnostic distribution shift in this paper. As we discussed above, the agnostic distribution shift can result in correlations between features. 
    
\subsection{Problem Definition}
    We first introduce a formal definition of the OOD image classification problem proposed by \cite{he2021towards} as follows: 
    \paragraph{OOD Image Classification} Given the training dataset \(\mathcal{D}_{train}=\{(x_{i}, y_{i})\}^{n_{train}}_{i=1}\) and the testing dataset $\mathcal{D}_{test}=\{(x_{i}, y_{i})\}^{n_{test}}_{i=1}$,  where $x_{i} \in \mathbb{R}^{c \times h \times w}$ represents an image sample, and $y_{i} \in \mathbb{R}^{1}$ represents its label, the task is to learn a feature extractor $f: \mathbb{R}^{c \times h \times w} \to \mathbb{R}^{p}$ with a dimension hyper-parameter $p$ and a classifier $z: \mathbb{R}^{p} \to \mathbb{R}^{1}$ so that $z\circ f$ can predict the labels of testing data precisely when data distributions $\psi(\mathcal{D}_{train}) \ne \psi(\mathcal{D}_{test})$, and $\psi(\mathcal{D}_{test})$ is unknown.
    
    \paragraph{Notations} In this work, capital letter refers to random vector and bold capital letter refers to matrix. $n$ refers to the sample size, and $p$ is the dimensions of variables. For any matrix $\mathbf{U}\in \mathbb{R}^{n \times p}$, we let $\mathbf{U}_{i,}$ and $\mathbf{U}_{,j}$ represent the $i$-th sample and the $j$-th variable in $\mathbf{U}$, respectively. $\mathbf{U}^{(-j)} \in \mathbb{R}^{n \times (p-1)}$ denotes the rest of $\mathbf{U}$ by removing its $j$-th dimension.

    \begin{figure*}[t]
        \centering
        \includegraphics[width=0.95\linewidth]{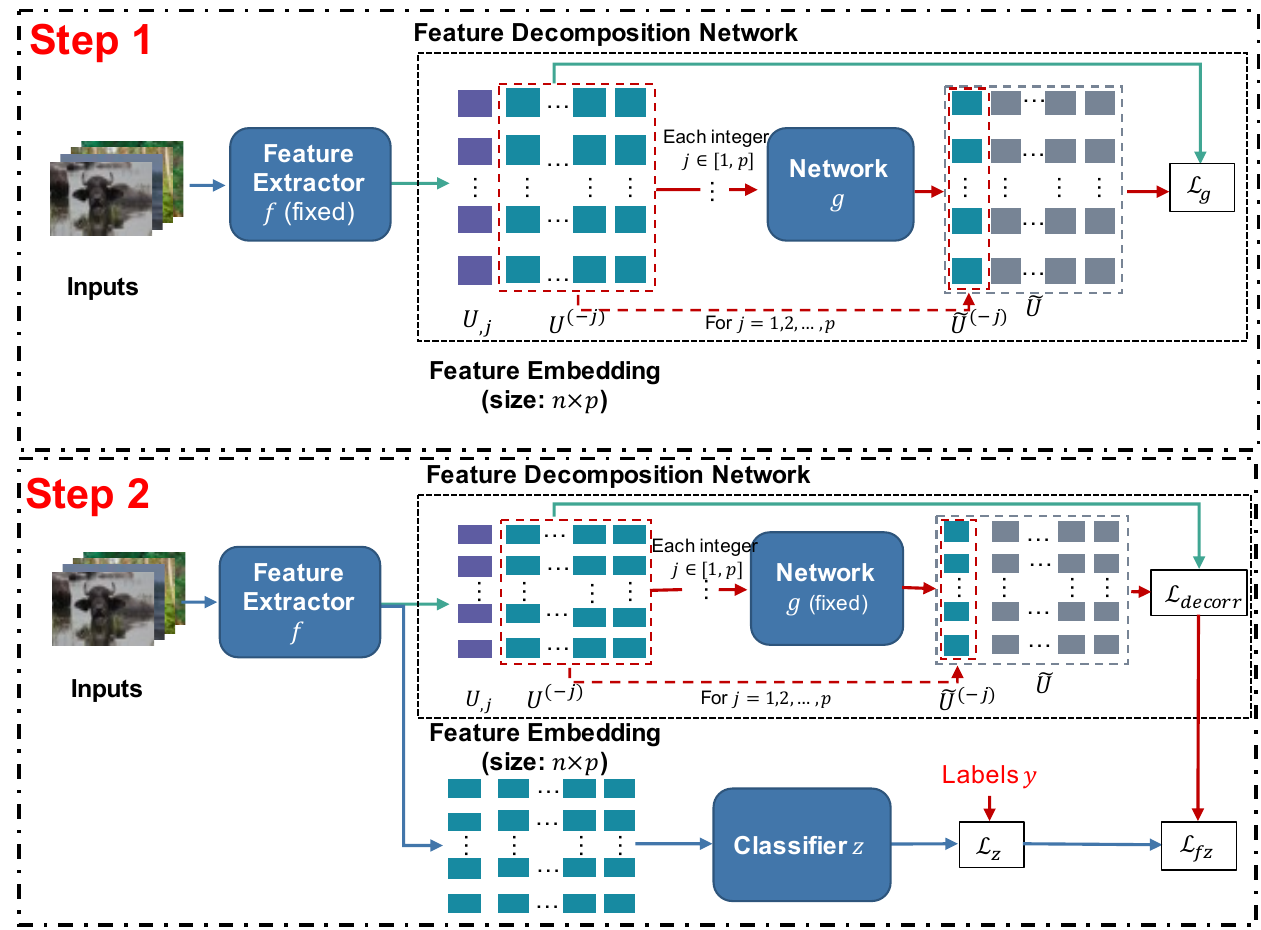}
        \caption{The framework of the proposed Partial Feature Decorrelation Learning (PFDL) algorithm. Step 1 and 2 are repeated for each mini-batch during training. $n$ denotes the sample size of each mini-batch. During inference, the feature decomposition network is dropped.}
        \label{fig:FDN}
    \end{figure*}

\subsection{Feature Decomposition Network}
    This subsection introduces our feature decomposition network, which decomposes feature embedding into the independent and correlated parts. 
    Let random vector $\mathrm{U}$ denotes the feature embedding extracted by $f$. Thus $\mathrm{U}$ represents feature embeddings of the training set $D_{train}$ extracted by $f$. We can treat $\mathrm{U} \in \mathbb{R}^{p}$ as a concatenation of random vectors $\mathrm{S}\in \mathbb{R}^{p_s}$ and $\mathrm{V} \in \mathbb{R}^{p_v}$, where $p = p_s + p_v$ and $\mathrm{U} = [\mathrm{S}^T; \mathrm{V}^T]^T$. Inspired by \cite{wang2006exploring}, we suppose $\mathrm{S}$ can be decomposed into two parts $\mathrm{S}=\mathrm{S}_{ind}+\hat{g}(\mathrm{V})$. The independent part $\mathrm{S}_{ind}$ is independent of $\mathrm{V}$, and the correlated part $\hat{g}(\mathrm{V})$ is a function of $\mathrm{V}$.
    
    We propose to learn the correlation function $\hat{g}:\mathbb{R}^{p_s} \to \mathbb{R}^{p_v} $ with a network $g:\mathbb{R}^{p_s} \to \mathbb{R}^{p_v} $. Then, we can divide $\mathbb{E}[\lVert \mathrm{S}-g(\mathrm{V})\rVert_{2}^{2}]$ as follows:
    
    \begin{equation}
    \begin{split}
        \mathbb{E}&[\lVert \mathrm{S}-g(\mathrm{V}) \rVert_{2}^{2}] = \mathbb{E}[\lVert \mathrm{S}_{ind}+\hat{g}(\mathrm{V})-g(\mathrm{V})\rVert_{2}^{2}] \\
        &= \mathbb{E}[\lVert (\mathrm{S}_{ind} - \mathbb{E}[\mathrm{S}_{ind}]) + (\hat{g}(\mathrm{V}) + \mathbb{E}[\mathrm{S}_{ind}]) - g(\mathrm{V})\rVert_{2}^{2}] \\
        &=\mathbb{E}[\lVert \mathrm{S}_{ind} - \mathbb{E}[\mathrm{S}_{ind}] \rVert_{2}^{2}] + \mathbb{E}[\lVert (\hat{g}(\mathrm{V}) + \mathbb{E}[\mathrm{S}_{ind}]) - g(\mathrm{V}) \rVert_{2}^{2}] \\
        &~~~~~~~~ + 2\,\mathbb{E}[\mathrm{S}_{ind} - \mathbb{E}[\mathrm{S}_{ind}]]^{T}\,\mathbb{E}[\hat{g}(\mathrm{V}) + \mathbb{E}[\mathrm{S}_{ind}] - g(\mathrm{V})] \\
        &=\mathbb{E}[\lVert \mathrm{S}_{ind} - \mathbb{E}[\mathrm{S}_{ind}] \rVert_{2}^{2}] + \mathbb{E}[\lVert (\hat{g}(\mathrm{V}) + \mathbb{E}[\mathrm{S}_{ind}]) - g(\mathrm{V}) \rVert_{2}^{2}]
    \end{split}
    \label{equ:e1}
    \end{equation}
    
    For simplicity but without loss of generality, we can assume that $\mathbb{E}[\mathrm{S}_{ind}]=0$, then \eqref{equ:e1} becomes:
    \begin{equation}
    \begin{split}
        \mathbb{E}[\lVert \mathrm{S}-g(\mathrm{V}) \rVert_{2}^{2}] =\mathbb{E}[\lVert \mathrm{S}_{ind} \rVert_{2}^{2}] + \mathbb{E}[\lVert \hat{g}(\mathrm{V}) - g(\mathrm{V}) \rVert_{2}^{2}]
    \end{split}
    \label{equ:e11}
    \end{equation}
    
    Since $\mathrm{S}_{ind}$ and $\mathrm{V}$ are independent, $\mathbb{E}[\lVert\mathrm{S}-g(\mathrm{V})\rVert^{2}_{2}]$ reaches minimum if and only if $g(\mathrm{V}) = \hat{g}(\mathrm{V})$.
    
    In conclusion, the solution to the optimization problem 
    \begin{equation}
    \min_{g} \mathbb{E}[\lVert \mathrm{S}-g(\mathrm{V}) \rVert_{2}^{2}]
     \label{equ:e2}
    \end{equation} 
    is $\hat{g}$. 
    
    %So we can learn the function $\hat{g}$ by optimizing:
    %\begin{equation}
    %    \min_{g} \frac{1}{n} \sum_{i=1}^{n} \lVert \mathbf{s_{i}} - g(\mathbf{v_{i}})\rVert^{2}_{2}.
    %\end{equation}

    Thanks to the universal approximation theorem \cite{hornik1989multilayer}, we can use multi-layer perceptrons (MLPs) to model the correlations between $\mathrm{S}$ and $\mathrm{V}$.
    
    In practice, a training set $D_{train}=\{(\mathbf{x}_{i}, y_{i})\}^{n}_{i=1}$, and feature embedding $\mathbf{U}_{i} = f(\mathbf{x}_{i})$ is given, where $\mathbf{U} \in \mathbb{R}^{n \times p}$ and $p$ denotes the dimension of the feature embedding. In our model, for each feature dimension, we successively consider each $\mathbf{U}_{,j} \in \mathbb{R}^{n \times 1}$ of the feature embedding as the random sample of random vector $\mathrm{S}$ and the rest part $\mathbf{U}^{(-j)} \in \mathbb{R}^{n \times (p-1)}$ as the random sample of random vector $\mathrm{V}$. Then we have the loss function of the feature decomposition network $g$ as follows:
    \begin{equation}
        \mathcal{L}_{g} = \frac{1}{n}\,\frac{1}{p}\,\sum_{j=1}^{p}\,\sum_{i=1}^{n}\,\lVert\,\mathbf{U}_{i,j} - \widetilde{\mathbf{U}}_{i,j} \rVert_{2}^{2}
        \label{equ:fdf}
    \end{equation}
    where $\widetilde{\mathbf{U}}_{i,j} = g(\mathbf{U}^{(-j)}_{i,})$, and $\mathbf{U}^{(-j)}_{i,}$ denotes the sample $\mathbf{x}_{i}$'s feature embedding except the $j$-th dimension.
    
    However, suppose $D_{train}$ and $D_{test}$ are generated under OOD generalization settings. In that case, the training set $D_{train}$ is biased from the true distribution. The model will inevitably overfit the correlations only existing in the training set and omit some causality and connections between features. It can make the residual $\widetilde{\mathrm{S}}_{ind}$ of the misspecified model deviate from the true $\mathrm{S}_{ind}$. 
    
    Hence, if we can adjust the feature extractor to reduce the distance between $\widetilde{\mathrm{S}}_{ind}$ and $\mathrm{S}_{ind}$, we can learn a stable image classification model. Although the true $\mathrm{S}_{ind}$ is unknown, we can still lower the loss in \eqref{equ:fdf} by iteratively optimizing $g$ and the target image classification model. There is no prior knowledge about which part of the feature embedding is the effect of correlation features in real-world applications. Hence, as shown in \eqref{equ:fdf} and Figure.~\ref{fig:FDN}~'Step 1', we successively modeling each dimension $\mathbf{U}_{,j}$ of the feature embedding using a parameter-shared feature decomposition network $g$.

\subsection{Partial Feature Decorrelation Algorithm}
    In this subsection, we introduce the partial feature decorrelation learning algorithm, which uses \eqref{equ:fdf} and $g$ to help learn a stable feature representation.
    
    Our goal is to decorrelate correlated features. Given a feature extractor $f$, we use the well-trained and fixed feature decomposition network $g$ to decompose the correlated part features when training feature extractor $f$ and classifier $z$. 
    
    In general cases, assume that $\mathbf{X} \in \mathbb{R}^{n \times p_x}$ and $\mathbf{Y} \in \mathbb{R}^{n \times p_y}$ are the random samples of random vectors $\mathrm{X} \in \mathbb{R}^{p_x}$ and $\mathrm{Y} \in \mathbb{R}^{p_x}$. If $\mathrm{X}$ and $\mathrm{Y}$ are independent, we have $\operatorname{cov}(\textbf{X},\textbf{Y}) = \mathbb{E}[\textbf{X}^T\textbf{Y}] - \mathbb{E}[\textbf{X}]^T\mathbb{E}[\textbf{Y}] = 0$. Thus, $\mathbb{E}[\textbf{X}^T\textbf{Y}] = \mathbb{E}[\textbf{X}]^T\mathbb{E}[\textbf{Y}]$. Moreover, if $\mathbb{E}[\mathbf{Y}]=0$, then $\mathbb{E}[\mathbf{X}^{T}\mathbf{Y}] = 0$. For simplicity but without loss of generality, we assume that $\mathbb{E}[\mathrm{V}] =0$. Hence, we propose the feature decorrelation constraint that minimize the correlation between $g(\mathrm{V})$ and $\mathrm{V}$ to partially decorrelate features, when training the feature extractor $f$. Formally:
    \begin{equation}
        \mathcal{L}_{decorr} =  \frac{1}{p}\,\sum_{j=1}^{p}\,\lVert\,\frac{1}{n(p-1)}\,\sum_{i=1}^{n}\sum_{k\neq j}\,\widetilde{\mathbf{U}}_{i,j} \mathbf{U}^{(-j)}_{i,k} \rVert^{2}_{2}
    \end{equation}
    where $\widetilde{\mathbf{U}}_{i,j} = g(\mathbf{U}^{(-j)}_{i,})$, in which the parameters of network $g$ is fixed and optimized in advance using \eqref{equ:fdf}. Because $\hat{g}(\mathrm{V})$ is the correlated part of $\mathrm{S}$, our setting in fact minimize the correlation between $\mathrm{S}$ and $\mathrm{V}$.
    
    The overall loss function for training $f$ and $z$ is:
    \begin{equation}
    \begin{split}
        \mathcal{L}_{fz} = \frac{1}{n}\,\sum_{i=1}^{n} \mathcal{L}_{n}(z(\mathbf{U}_{i,})) + \mathcal{L}_{decorr}
    \end{split}
    \label{equ:pdt}
    \end{equation}
    where $\mathcal{L}_{n}$ is the original image classification loss function, \textit{e.g.,} a Cross-Entropy loss function. 
    
    Based on \eqref{equ:fdf} and \eqref{equ:pdt}, we develop the Partial Feature Decorrelation Learning (PFDL) algorithm for OOD image classification, shown in Algorithm.~\ref{alg:algorithm}. The flow chart of the proposed PFDL algorithm has been shown in Figure.~\ref{fig:FDN}. 
    
    As shown in Figure.~\ref{fig:FDN} and Algorithm.~\ref{alg:algorithm}, for each mini-batch, we first train the feature decomposition network in 'Step 1' with the fixed feature extractor using \eqref{equ:fdf}. We then train the feature extractor and classifier in 'Step 2' with the fixed feature decomposition network using \eqref{equ:pdt}.
    
    \paragraph{Connections to Causal Inference}
    Causal effect decomposition methods \cite{wang2006exploring,wang2008casual} shed some lights on our model. The causal effect decomposition method is intended to partition observed associations among variables into direct and indirect effects when independent relationships exist \cite{wang2008casual}. By doing so, one can better analyze the direct effects of treatment features on the outcome. In fact, the direct and indirect effects are connected to the independent and correlated parts in our feature decomposition network.
    
    \paragraph{Connections to Information Theory}
    From \cite{paninski2003estimation}, we know that two random vectors $\mathrm{X}$ and $\mathrm{Y}$ with values over the space $\mathcal{X} \times \mathcal{Y}$ are independent if and only if the mutual information between $\mathrm{X}$ and $\mathrm{Y}$ is equal to 0. In fact, our motivation can also be interpreted as learning a discriminative feature representation by reducing the mutual information between the independent part features and other features. 
    
    \begin{algorithm}
    \caption{Partial Feature Decorrelation Learning}
    \label{alg:algorithm}
        \textbf{Input}: Training set $D_{train}=\{(\mathbf{x}_{i}, y_{i})\}^{n}_{i=1}$, maximum epoch number $E$, initialized feature extractor $f^{(0)}$, classifier $z^{(0)}$ and feature decomposition network $g^{(0)}$ \\
        \textbf{Output}: Optimized $f$, $z$
        \begin{algorithmic}[1] %[1] enables line numbers
        \STATE $t=0$
        \FOR{$e$ in range($0, E$)}
        \REPEAT
        \STATE Sample mini-batch $\{\mathbf{x},y\}_{t}$ from $D_{train}$
        \STATE Calculate $\mathcal{L}_{g}$ using \eqref{equ:fdf} with ($g^{(t)}$, $f^{(t)}$, $\{\mathbf{x},y\}_{t}$)
        \STATE Update $g^{(t+1)}$ with a stochastic gradient descent optimizer by fixing $f^{(t)}$
        \STATE Calculate $\mathcal{L}_{fz}$ using \eqref{equ:pdt} with ($g^{(t+1)}$, $f^{(t)}$, $z^{(t)}$, $\{\mathbf{x},y\}_{t}$)
        \STATE Update $f^{(t+1)}, z^{(t+1)}$ with a stochastic gradient descent optimizer by fixing $g^{(t+1)}$
        \STATE $t = t+1$
        \UNTIL $D_{train}$ has been traversed
        \ENDFOR
        \RETURN $f^{(t)}$ and $z^{(t)}$
    \end{algorithmic}
    \end{algorithm}

\section{Experiments}
\label{sec:exp}
    We evaluate the proposed PFDL algorithm in terms of the model performance quantitatively and qualitatively. We first introduce the datasets and evaluation metrics. We verify the correlation modeling ability of the proposed Feature Decomposition Network on a synthetic OOD prediction dataset MMADS \cite{kuang2020stable} compared with several baseline methods and a state-of-the-art stable learning method \cite{kuang2020stable}. We then demonstrate the effectiveness of PFDL on a OOD image classification dataset NICO \cite{he2021towards}, comparing it with several baseline and state-of-the-art methods.

\subsection{Datasets}
    \paragraph{MMADS} This synthetic dataset is to mimic the agnost distribution shift problem in which part of the causality is omitted due to the data generation bias, resulting in a statistical correlations between features. The task on this dataset is to predict the ground-truth using the input features. Given a input feature vector $\mathrm{X} = [\mathrm{S}^{T};\mathrm{V}^{T}]^{T}$ (where $\mathrm{X} \in \mathbb{R}^{p}$, $\mathrm{S}\in \mathbb{R}^{p_{s}}$, $\mathrm{V} \in \mathbb{R}^{v}$, and $p = p_s + p_v$) and the ground-truth $\mathrm{Y} \in \mathbb{R}^{1}$, the stable feature $\mathrm{S}$ has real causality with the ground-truth $\mathrm{Y}$, and correlation feature $\mathrm{V}$ does not. 
    Following the setting used by the state-of-the-art stable learning method DWR \cite{kuang2020stable}, we mimic three kinds of relationship between $\mathrm{S}$ and $\mathrm{V}$, including 'independent' ($\mathrm{S}\bot \mathrm{V}$), '$\mathrm{V}$ depends on $\mathrm{S}$' ($\mathrm{S}\rightarrow \mathrm{V}$) and '$\mathrm{S}$ depends on $\mathrm{V}$' ($\mathrm{S}\leftarrow \mathrm{V}$). 
    
    $\mathrm{S}\bot \mathrm{V}$: In this setting, $\mathrm{S}$ and $\mathrm{V}$ are independent, but $\mathrm{S}_{,i}$ could be dependent with each other. Hence, we generate $\mathrm{X} = \{\mathrm{S}_{,1},...,\mathrm{S}_{,p_s},\mathrm{V}_{,1},...,\mathrm{V}_{,p_v}\}$ with independent Gaussian distributions with the help of auxiliary variables $\mathrm{Z}$ as following:
        \begin{equation}
        \label{equ:zv_gen}
            \mathrm{Z}_{,1}, ..., \mathrm{Z}_{,p} \stackrel{\text{IID}}{\sim} \mathcal{N}(0,1); \mathrm{V}_{,1}, ..., \mathrm{V}_{,p_v} \stackrel{\text{IID}}{\sim} \mathcal{N}(0,1) 
        \end{equation}
        \begin{equation}
        \label{equ:s_gen}
            \mathrm{S}_{,i} =0.8*\mathrm{Z}_{,i}+0.2*\mathrm{Z}_{,i+1}, i=1,2,...,p_s
        \end{equation}
        where the number of stable variables $p_s = 0.5 * p$ and the number of unstable variables $p_v = 0.5 * p$. $\mathrm{S}_{,j}$ represents the $j$-th variable in $\mathrm{S}$.
        
    $\mathrm{S}\rightarrow \mathrm{V}$: In this setting, the stable features $\mathrm{S}$ are the causes of unstable features $\mathrm{V}$. We first generate dependent stable features $\mathrm{S}$ with Eq.~\eqref{equ:s_gen}. Then, we generate unstable features $\mathrm{V}$ based on $\mathrm{S}$: $\mathrm{V}_{,j}=0.8*\mathrm{S}_{,j}+0.2*\mathrm{S}_{,j+1}+\mathcal{N}(0,1)$, where we let $j+1 = mod(j+1, p_s)$. The function $mod(a, b)$ returns the modulus after division of $a$ by $b$.
    
    $\mathrm{S}\leftarrow \mathrm{V}$: In this setting, unstable features $\mathrm{V}$ are the causes of stable features $\mathrm{S}$. We first generate the unstable features $\mathrm{V}$ with Eq.~\eqref{equ:zv_gen}. Then, we generate the stable features $\mathrm{S}$ based on $\mathrm{V}$: $\mathrm{S}_{,j} =0.2*\mathrm{V}_{,j} +0.8*\mathrm{V}_{,j+1}+\mathcal{N}(0,1)$, where we let $j+1=mod(j+1,p_v)$.

    The relationship among $\mathrm{S}$, $\mathrm{V}$ and $\mathrm{Y}$ has the form: $\mathrm{Y}_{poly} = [\mathrm{S}, \mathrm{V}] \cdot [\beta_{\mathrm{S}}, \beta_{\mathrm{V}}]^{T} + \mathrm{S}_{,1}\,\mathrm{S}_{,2}\,\mathrm{S}_{,3}+\epsilon$, where $\mathrm{Y}_{poly}$ denotes the ground-truth generated from a polynomial nonlinear function, $\beta_{\mathrm{S}} = \{\frac{1}{3},-\frac{2}{3},1,-\frac{1}{3},\frac{2}{3},-1,...\}$, $\beta_{\mathrm{V}} = \Vec{0}$ and $\epsilon = \mathcal{N}(0,0.3)$.
    
    To mimic the OOD generalization settings, we generate a set of environments, each with a distinct joint distribution $P^{e}(\mathrm{X},\mathrm{Y})$ while preserving $P(\mathrm{Y}|\mathrm{S})$ as the same. To achieve that, we generate environments by varying $P^{e}(\mathrm{V}_{b}|\mathrm{S})$ on a subset $\mathrm{V}_{b} \in \mathrm{V}$. Following the setting used by DWR \cite{kuang2020stable}, we vary $P^{e}(\mathrm{V}_{b}|\mathrm{S})$ via biased sample selection with a bias rate $r \in [-3,-1) \cup (1,3]$. For each sample, the probability of being selected is defined as $P^{e}_{r} = \prod_{\mathrm{V}_{i} \in \mathrm{V}_{b}} |r|^{-5*D_{i}}$, where $D_{i} = |f(\mathrm{S}) - sign(r)*\mathrm{V}_{i}|$. If $r>0, sign(r)=1$, otherwise $sign(r)=-1$.
    
    In this paper, we choose the number of sample size $n$ from $\{1000, 2000, 4000\}$, the feature dimension $p$ from $\{10, 20, 40\}$ and the bias rate from $\{1.5, 1.7, 2.0\}$. The details of the discussion and results are given in Sec.~\ref{sec:mmads-results}.
    
    \paragraph{NICO} NICO \cite{he2021towards} dataset is a OOD image classification dataset, which contains 25,000 images obtained from the Internet. NICO contains two superclasses: Animal and Vehicle, with 10 classes for Animal and 9 classes for the Vehicle. Specifically, there are 9 or 10 contexts in each class, and each image has been manually annotated in its context. The average size of classes is about 1300 images. 
    
    We evaluate our model on the NICO dataset using the Animal superclass. We used the proportional bias setting used by previous work \cite{he2021towards}, in which all contexts are used on both training and testing sets, but the percentage of each context is different on training and testing sets. Particularly, one context of each class will be randomly selected as the dominant context for both training and testing, and the rest is considered minor contexts. The dominant ratio \cite{he2021towards} is defined as $Dominant~ratio = N_{dominant}/N_{minor}$ where $N_{dominant}$ refers to the sample size of the dominant context and $N_{minor}$ refers to the average size of other contexts where we uniformly sample other contexts.
    
    \paragraph{Colored-MNIST} This dataset is a synthetic dataset formed by coloring each image either red or green in a way that correlates strongly (but spuriously) with the class label. We followed the experiment setting of state-of-the-art work IRM \cite{arjovsky2020invariant}.

%\subsection{Implementation Details}
%    For MMADS, the backbone of feature extractor is, the classifier is , the feature decomposition networks is ..
%    For NICO,
%    For Colored-MNIST, 
    
%    The learning rate, optimization method, random initialization...
    
%    (-) The proposed model is not clearly written. The proposed model consists of feature extractor f() and classifier z(), feature decomposition networks g(). The architectures of these modes are not written.

%    (-) How to apply the proposed method to artificial data is not clear. The feature extractor of the proposed method and classifier seems to be not used in the MMADS dataset.

\subsection{Evaluation Metrics}
    \paragraph{MMADS} We use coefficient estimation error $\beta\text{\_}error$ \cite{kuang2020stable} to evaluate the correlation modeling ability of the feature decomposition network. The $\beta\text{\_}error$ between the learned coefficient $\hat{\beta}$ and the true coefficient $\beta$ is defined as $\beta\text{\_}error = \frac{1}{p}\|\beta - \hat{\beta}\|_1$, where $p$ is the feature dimension of $\beta$. We report both mean and variance of $\beta\text{\_}error$ of five independent experiments. Moreover, we use RMSE, Average Error (AE) \cite{kuang2020stable} and Stability Error (SE) \cite{kuang2020stable} to evaluate stability of our model. 
    
    \paragraph{NICO and Colored-MNIST} Following \cite{arjovsky2020invariant,he2021towards}, we use accuracy to evaluate the classification ability in both NICO and Colored-MNIST datasets.

\subsection{Experiments on MMADS Dataset}
\label{sec:mmads-results}
    \begin{figure*}[t]
      \centering
      \subfloat[$\mathrm{S}\bot \mathrm{V}$]{\includegraphics[width=\textwidth]{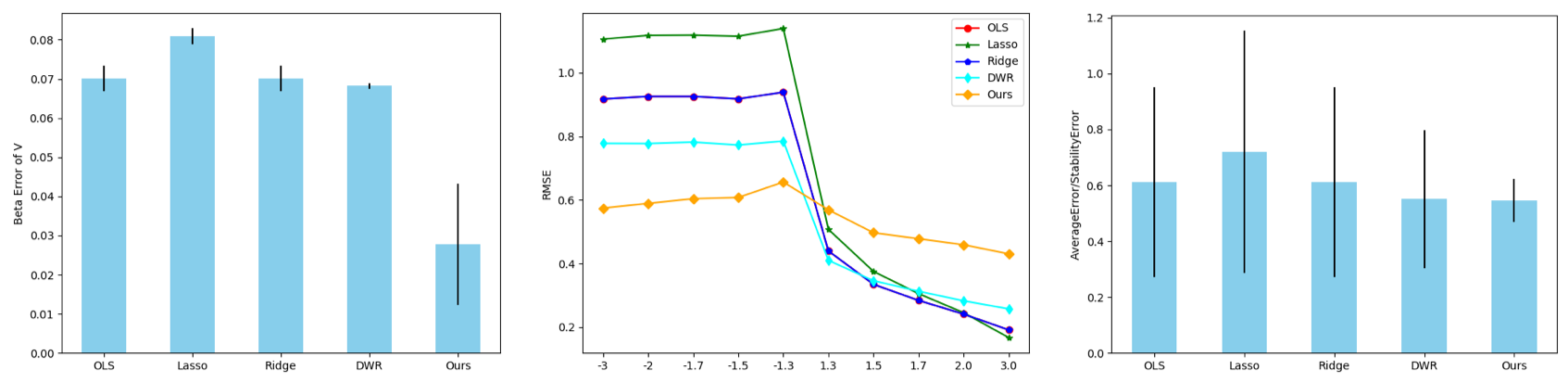}} \\
      \subfloat[$\mathrm{S}\leftarrow \mathrm{V}$]{\includegraphics[width=\textwidth]{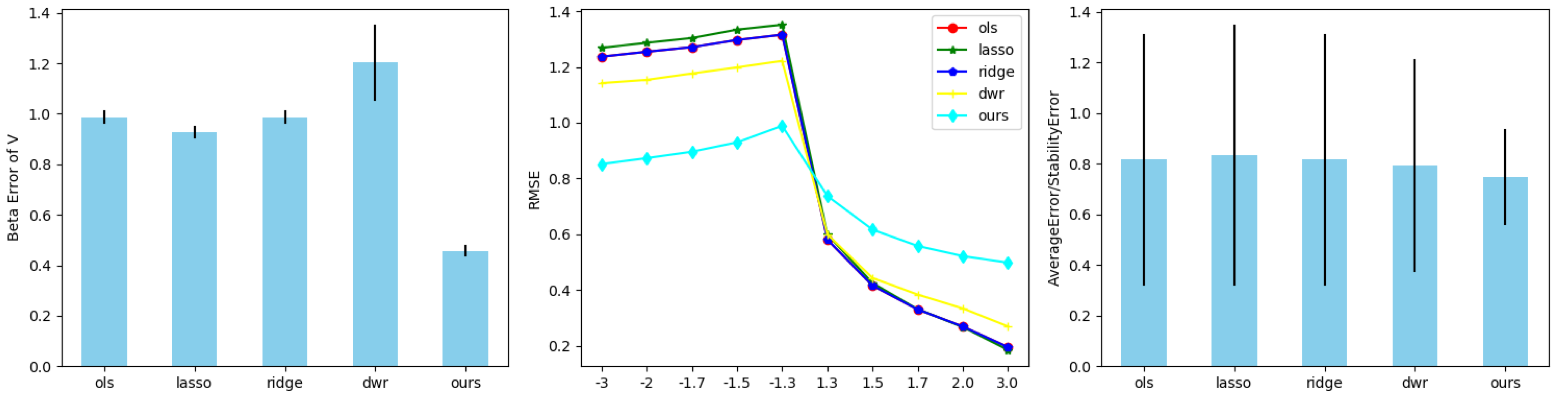}} \\
      \subfloat[$\mathrm{S}\rightarrow \mathrm{V}$]{\includegraphics[width=\textwidth]{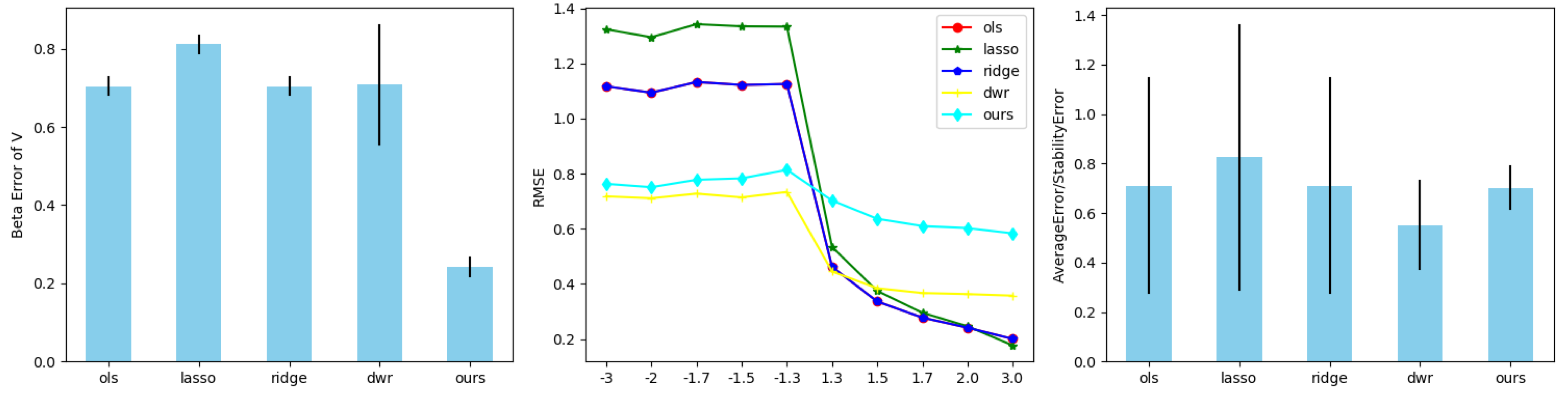}}
      \caption{Experimental results with causality setting $\mathrm{S}\bot \mathrm{V}$ and nonlinear function $\mathrm{Y} = \mathrm{Y}_{poly}$. All models are trained with $n=2000, p=20, r_{train}=1.7$.}
      \label{fig:sindv}
    \end{figure*}
    
    \begin{table*}[t]
    \caption{Experimental results under setting $\mathrm{S}\bot \mathrm{V}$ with $\mathrm{Y} = \mathrm{Y}_{poly}$ when varying sample size $n$, feature dimension $p$ and training bias rate $r$. The smaller value in this table, the better. We use bold font to highlight the results of our model.}
    \label{tab:sindv}
    \resizebox{\linewidth}{!}{
    \begin{tabular}{cccccccccccccccc}
        \hline
        \multicolumn{16}{c}{Scenario 1: varying sample size n}                                                                                                                          \\ \hline
        n,p,r          & \multicolumn{5}{|c|}{n=1000,p=10,r=1.7}                     & \multicolumn{5}{c|}{n=2000,p=10,r=1.7}                     & \multicolumn{5}{c}{n=4000,p=10,r=1.7} \\ \hline
        Methods        & \multicolumn{1}{|c}{OLS}   & Lasso & Ridge & DWR   & \multicolumn{1}{c|}{Ours}   & OLS   & Lasso & Ridge & DWR   & \multicolumn{1}{c|}{Ours}   & OLS   & Lasso & Ridge & DWR   & Ours   \\ \hline
        %$\beta\text{\_s\_error}$ & \multicolumn{1}{|c}{0.260} & 0.313 & 0.260 & 0.189 & \multicolumn{1}{c|}{\textbf{0.369}} & 0.262 & 0.315 & 0.262 & 0.192 & \multicolumn{1}{c|}{\textbf{0.371}} & 0.264 & 0.317 & 0.264 & 0.203 & \textbf{0.453} \\
        $\beta_\mathrm{V}\text{\_error}$ & \multicolumn{1}{|c}{0.099} & 0.102 & 0.099 & 0.066 & \multicolumn{1}{c|}{\textbf{0.027}} & 0.097 & 0.101 & 0.097 & 0.060 & \multicolumn{1}{c|}{\textbf{0.025}} & 0.097 & 0.101 & 0.097 & 0.057 & \textbf{0.016} \\
        AE             & \multicolumn{1}{|c}{0.604} & 0.639 & 0.603 & 0.519 & \multicolumn{1}{c|}{\textbf{0.629}} & 0.583 & 0.617 & 0.583 & 0.509 & \multicolumn{1}{c|}{\textbf{0.613}} & 0.587 & 0.621 & 0.587 & 0.505 & \textbf{0.569} \\
        SE             & \multicolumn{1}{|c}{0.254} & 0.285 & 0.254 & 0.103 & \multicolumn{1}{c|}{\textbf{0.086}} & 0.236 & 0.267 & 0.236 & 0.110 & \multicolumn{1}{c|}{\textbf{0.071}} & 0.236 & 0.267 & 0.236 & 0.114 & \textbf{0.089} \\ \hline
        \multicolumn{16}{c}{Scenario 2: varying feature dimension p}                                                                                                                     \\ \hline
        n,p,r          & \multicolumn{5}{|c|}{n=2000,p=10,r=1.7}                     & \multicolumn{5}{c|}{n=2000,p=20,r=1.7}                     & \multicolumn{5}{c}{n=2000,p=40,r=1.7} \\ \hline
        Methods        & \multicolumn{1}{|c}{OLS}   & Lasso & Ridge & DWR   & \multicolumn{1}{c|}{Ours}   & OLS   & Lasso & Ridge & DWR   & \multicolumn{1}{c|}{Ours}   & OLS   & Lasso & Ridge & DWR   & Ours   \\ \hline
        %$\beta\text{\_s\_error}$ & \multicolumn{1}{|c}{0.262} & 0.315 & 0.262 & 0.192 & \multicolumn{1}{c|}{\textbf{0.371}} & 0.339 & 0.451 & 0.339 & 0.282 & \multicolumn{1}{c|}{\textbf{0.455}} & 0.892 & 0.907 & 0.907 & 0.912 & \textbf{0.578} \\
        $\beta_\mathrm{V}\text{\_error}$ & \multicolumn{1}{|c}{0.097} & 0.101 & 0.097 & 0.060 & \multicolumn{1}{c|}{\textbf{0.025}} &  0.070 & 0.080 & 0.070 & 0.066 & \multicolumn{1}{c|}{\textbf{0.027}} & 0.044 & 0.047 & 0.044 & 0.038 & \textbf{0.013} \\
        AE             & \multicolumn{1}{|c}{0.583} & 0.617 & 0.583 & 0.509 & \multicolumn{1}{c|}{\textbf{0.613}} &  0.612 & 0.720 & 0.612 & 0.550 & \multicolumn{1}{c|}{\textbf{0.546}} & 0.538 & 0.618 & 0.538 & 0.519 & \textbf{0.471} \\
        SE             & \multicolumn{1}{|c}{0.236} & 0.267 & 0.236 & 0.110 & \multicolumn{1}{c|}{\textbf{0.071}} & 0.319 & 0.408 & 0.319 & 0.232 & \multicolumn{1}{c|}{\textbf{0.071}} & 0.312 & 0.370 & 0.312 & 0.297 & \textbf{0.082} \\ \hline
        \multicolumn{16}{c}{Scenario 3: varying bias rate r on training data}                                                                                                            \\ \hline
        n,p,r          & \multicolumn{5}{|c|}{n=2000,p=20,r=1.5}                     & \multicolumn{5}{c|}{n=2000,p=20,r=1.7}                     & \multicolumn{5}{c}{n=2000,p=20,r=2.0} \\ \hline
        Methods        & \multicolumn{1}{|c}{OLS}   & Lasso & Ridge & DWR   & \multicolumn{1}{c|}{Ours}   & OLS   & Lasso & Ridge & DWR   & \multicolumn{1}{c|}{Ours}   & OLS   & Lasso & Ridge & DWR   & Ours   \\ \hline
        %$\beta\text{\_s\_error}$ & \multicolumn{1}{|c}{0.339} & 0.451 & 0.339 & 0.282 & \multicolumn{1}{c|}{\textbf{0.387}} & 0.339 & 0.451 & 0.339 & 0.282 & \multicolumn{1}{c|}{\textbf{0.455}} & 0.473 & 0.610 & 0.473 & 0.415 & \textbf{0.512} \\
        $\beta_\mathrm{V}\text{\_error}$ & \multicolumn{1}{|c}{0.059} & 0.067 & 0.059 & 0.060 & \multicolumn{1}{c|}{\textbf{0.010}} & 0.070 & 0.080 & 0.070 & 0.066 & \multicolumn{1}{c|}{\textbf{0.027}} & 0.079 & 0.091 & 0.079 & 0.077 & \textbf{0.023} \\
        AE             & \multicolumn{1}{|c}{0.519} & 0.590 & 0.519 & 0.548 & \multicolumn{1}{c|}{\textbf{0.497}} & 0.612 & 0.720 & 0.612 & 0.550 & \multicolumn{1}{c|}{\textbf{0.546}} & 0.660 & 0.781 & 0.660 & 0.613 & \textbf{0.618} \\
        SE             & \multicolumn{1}{|c}{0.220} & 0.297 & 0.220 & 0.197 & \multicolumn{1}{c|}{\textbf{0.031}} & 0.319 & 0.408 & 0.319 & 0.232 & \multicolumn{1}{c|}{\textbf{0.071}} & 0.364 & 0.447 & 0.364 & 0.303 & \textbf{0.119} \\ \hline
    \end{tabular}}
    \end{table*}
    
    We first use the synthetic dataset MMADS to quantitatively and qualitatively evaluate the decorrelation ability of our model since all correlations are manually controlled. To mimic the OOD generalization scenario, we use different probability $P(\mathrm{V}_{b}|\mathrm{S})$ on a subset $\mathrm{V}_{b}\in \mathrm{V}$ for training and testing environment to create correlations between $\mathrm{V}$ and the ground-truth $\mathrm{Y}$ while simulating the agnostic distribution shift between training and testing. By doing so, we can use the coefficient estimation error on correlation features $\mathrm{V}$ as a quantitative index to evaluate how our method models correlations. We also use Average Error (AE) and Stability Error (SE) to evaluate the stability improvement provided by our method. 
    
    \paragraph{Compared Methods}
    We use four methods as baselines on the MMADS dataset, including the state-of-the-art stable learning method DWR \cite{kuang2020stable} and three IID-based machine learning baseline methods OLS~\cite{hutcheson2011ordinary}, Lasso \cite{tibshirani1996regression}, and Ridge Regression \cite{hoerl1970ridge}. We used the official implementation of DWR provided by the authors.
    
    \paragraph{Implementation}
    All models are trained on the same training dataset generated using a specific bias rate $r_{train}$. We repeat this training process five times with different $r_{train}$ and report the mean and variance of $\beta\text{\_}error$ on $\mathrm{V}$, since the $\mathrm{V}$ is correlation features correlated with $\mathrm{Y}$ due to data generation bias. To evaluate the prediction stability, we test all models on several test environments with various bias rates $r_{test} \in  [-3,-1) \cup (1,3]$. For each test bias rate, five different test datasets are generated. To verify the correlation modeling ability of our method, we directly use the output of the feature decomposition network as auxiliary features, feed it into the coefficient estimator (same as DWR) to fit the biased training data, and then evaluate the coefficient estimator error.
    
    For DWR, we used the hyper-parameters provided in their official implementation. The learning rate of DWR is 0.005, and the maximum iteration is 5000. As for the hyper-parameters of PFDL, the learning rate is 0.001, and the maximum iteration is the same as DWR.
    
    As for computing resources, all experiments on the MMADS dataset are completed with a MacBook Pro laptop without GPU acceleration. Each experiment on the MMADS dataset is completed in minutes, including the training and testing process.
    
    \paragraph{Results}
    The experimental results are shown in Figure.~\ref{fig:sindv} and Table.~\ref{tab:sindv}. we visualized the results of three different correlation settings $\mathrm{S}\bot \mathrm{V}, \mathrm{S}\leftarrow \mathrm{V}, \mathrm{S}\rightarrow \mathrm{V}$. We can notice that our algorithm can achieve the lowest coefficient estimation error on $\beta_\mathrm{V}$ compared with all baselines. It shows that our method can model the correlations between features accurately in all three different correlations settings. Similar evidence can be found in Table.~\ref{tab:sindv}, where our method achieved the lowest $\beta_\mathrm{V}\text{\_error}$ and SE in all experiments compared with all baselines, including state-of-the-art method DWR. Besides, we can notice that our method can significantly mitigate the model misspecification caused by correlations and improve the stability of the model across different test datasets by using the output of the feature decomposition network. 
    
    We can notice that the performance of our proposal is stable when the sample size changing, but the performance of the state-of-the-art method DWR is affected by the sample size as shown in Table \ref{tab:sindv}. It shows that our method is more robust when training with a small number of samples.

    These observations lead to the conclusion that our proposal can achieve better stability across different test datasets than the state-of-the-art methods.

\subsection{Experiments on NICO Dataset}

    \begin{table*}[!ht]
        \centering
        \caption{Results on NICO dataset. The evaluation metric is standard classification accuracy in percentage. Models are trained on the same training set with $Dominant~Ratio = 5:1$. The symbol '-' denotes that the previous work has not implemented this setting. We added these settings to show a fine-grained evaluation.}
        \label{tab:nico1}
        \begin{tabular}{lccccccc}
        \hline
        $Dominant~Ratio$& $1:5$ & $1:3$ & $1:1$ & $2:1$ & $3:1$ & $4:1$ & $5:1$   \\ \hline
        DWR \cite{kuang2020stable} & 35.66 &35.61 & 37.29 & 40.37 & 41.81 & 43.08 &43.72 \\
        CN \cite{he2021towards} & 37.17 & - & 37.80 & 41.46 & 42.50 & 43.23 & -  \\
        CN+BN \cite{he2021towards}   & 38.70 & - & 39.60 & 41.64 & 42.00 & 43.85 & -       \\
        CNBB \cite{he2021towards} & 39.06 & - & 39.60 & 42.12 & 43.33 & 44.15 & -             \\  \hline
        %ResNet-50 \cite{he2016deep}   & 39.79 & 39.98 & 40.54 & 42.41 & 43.21 & 46.67 & 44.73       \\ 
        PFDL (Ours)   & \textbf{41.31}  & \textbf{41.41} & \textbf{43.70} & \textbf{47.46} & \textbf{46.80} & \textbf{49.29} & \textbf{50.87}       \\ 
        \hline  
        \end{tabular}
    \end{table*} 
    
    We then demonstrate the effectiveness of our model on the real-world OOD image classification dataset NICO. As discussed above, we use the proportional bias setting to show the stability improvement brought by our model compared with several baseline methods.
    
    \paragraph{Compared Methods} We compare our model with several state-of-the-art methods on NICO datasets, including CN, CN+BN, and CNBB \cite{he2021towards}. Moreover, we reproduced the state-of-the-art stable learning method DWR on the NICO dataset. The results of CN, CN+BN, and CNBB models are reported by \cite{he2021towards}. The backbone model CN is a typical ConvNet based neural network. The CN+BN denotes a typical CNN-based model with batch normalization. And the CNBB is the state-of-the-art method proposed by \cite{he2021towards}.
    
    \paragraph{Implementation}
    We then randomly sample $60\%$ images of each context of each class to form the training set and regard the rest as the testing set. In each class, we randomly sample a dominant context, and the rest becomes the minor contexts. For a dataset with $Dominant~Ratio = 1 : 5$, we will sample five minor context images of each minor context while we sample one dominant context image, until either dominant or minor context image runs out. Moreover, the context information is not available to the model during the training stage.
    
    In Table.~\ref{tab:nico1}, all models are trained on the same training set with $Dominant~Ratio = 5 : 1$, and tested on seven different testing sets with different $Dominant~Ratio$.
    
    As for hyper-parameters, the initial learning rate is 0.01 for all models and multiplied by 0.1 for every 80 epochs. All models are trained for 160 epochs. % We randomly select random seed 17 for all experiments.
    
    As for computing resources, all experiments on the NICO dataset are completed with a single Tesla P100 GPU. Each experiment is completed in hours, including the training and testing process. The computing performance demonstrates that our model does not consume too much extra computing resources.
    
\subsubsection{Results}
    Table.~\ref{tab:nico1} lists the OOD image classification performance on the NICO dataset. We can observe that PFDL deployed on a ResNet-50\cite{he2016deep} backbone achieves the best accuracy in all seven settings. We can notice that the accuracy of each model varies proportionally to the OOD generalization settings. Both our model and the baselines achieve the best performance in $Dominant~Ratio=5:1$, because all models are trained on the training set with $Dominant~Ratio=5:1$. Besides, we can notice that our model can significantly improve the classification accuracy in the most challenging setting $1:5$, demonstrating our model's effectiveness.

\subsection{Experiments on Colored-MNIST.}
    \begin{table}[t]
        \centering        
        \label{tab:colored-mnist}
        \caption{Results in Colored-MNIST.}
        \begin{tabular}{lcc}
        \hline
             Algorthm & Acc. train envs & Acc. test envs  \\
             \iffalse
             Random guessing (hypothetical) \cite{arjovsky2020invariant} & 50 & 50 \\
             Optimal invariant model (hypothetical) \cite{arjovsky2020invariant} &75 & 75 \\
             ERM, grayscale model (oracle) \cite{arjovsky2020invariant} & 73.5 $\pm$ 0.2 & 73.0$\pm$0.4 \\
             \fi
             ERM \cite{arjovsky2020invariant} & 87.40 $\pm$ 0.2 & 17.1 $\pm$ 0.6 \\
             IRM \cite{arjovsky2020invariant} & 70.8$\pm$0.9 & 66.9$\pm$2.5  \\ \hline
             IRM+Ours & \textbf{70.6}$\pm$\textbf{0.8} & \textbf{67.5} $\pm$\textbf{1.8} \\   \hline
        \end{tabular}
    \end{table}
    
    We then evaluate our model in the Colored-MNIST. The Colored-MNIST is a synthetic image classification dataset formed by coloring each image of the MNIST \cite{lecun1998gradient} dataset either red or green in a way that correlates strongly with the class label.
    
    \paragraph{Compared Methods} We compared our method with an MLP based model trained by using standard Empirical Risk Minimization (ERM) scheme. We also compared it with the state-of-the-art stable learning method IRM \cite{arjovsky2020invariant}.
    
    \paragraph{Implementation} We used the same implementation settings as in IRM \cite{arjovsky2020invariant}. The IRM model is also an MLP based neural network. We generate two training sets and one testing set with different correlation ratios (same as IRM \cite{arjovsky2020invariant}). The init learning rate is 0.001 for all methods. 
    
    \paragraph{Results} Table.~\ref{tab:colored-mnist} lists the image classification accuracy of all models. As we can notice, our model deployed in the IRM learning scheme can further improve the accuracy in test environments. It also demonstrates the effectiveness of our model.

\section{Conclusion}
    In this work, we study the OOD image classification problem. We present a novel Partial Feature Decorrelation Learning (PFDL) algorithm to improve the backbone model's stability by partially decorrelating features during training. Experiments on both synthetic and real-world datasets demonstrate that our method can help improve the stability of the backbone models and outperforms the state-of-the-art stable learning methods. We first verify the correlation modeling ability of our method on a synthetic dataset. We then demonstrate the effectiveness of our model on a OOD image classification dataset. 

\section*{Broader Impact}
    Our model is a data pretreatment method that causes no explicit ethical problems. As for benefits, our method can help improve the model's stability on unknown test datasets and reduce the data annotation costs for fine-tuning the model into new scenarios. It can be applied to most deep-learning-based and linear models in many application scenarios like quantitative financial analysis, social media, and time series-based forecasting tasks. Moreover, our method helps to leverage biases in the data. Thus the likely beneficiaries are companies that have business suffering the model misspecification problem. The consequence of the failure of the model is poor classification or regression performance. Our work may affect the interests of companies whose primary business is data annotations.

\ifCLASSOPTIONcompsoc
  \section*{Acknowledgments}
\else
  \section*{Acknowledgment}
\fi
This work was supported in part by The National Key Research and Development Program of China (Grant Nos: 2018AAA0101400), in part by The National Nature Science Foundation of China (Grant Nos: 62036009, U1909203, 61936006, 62133013), in part by Innovation Capability Support Program of Shaanxi (Program No. 2021TD-05).

\bibliographystyle{IEEEtran}
\bibliography{reference}

\newpage

\begin{IEEEbiography}
[{\includegraphics[width=1in,height=1.25in]{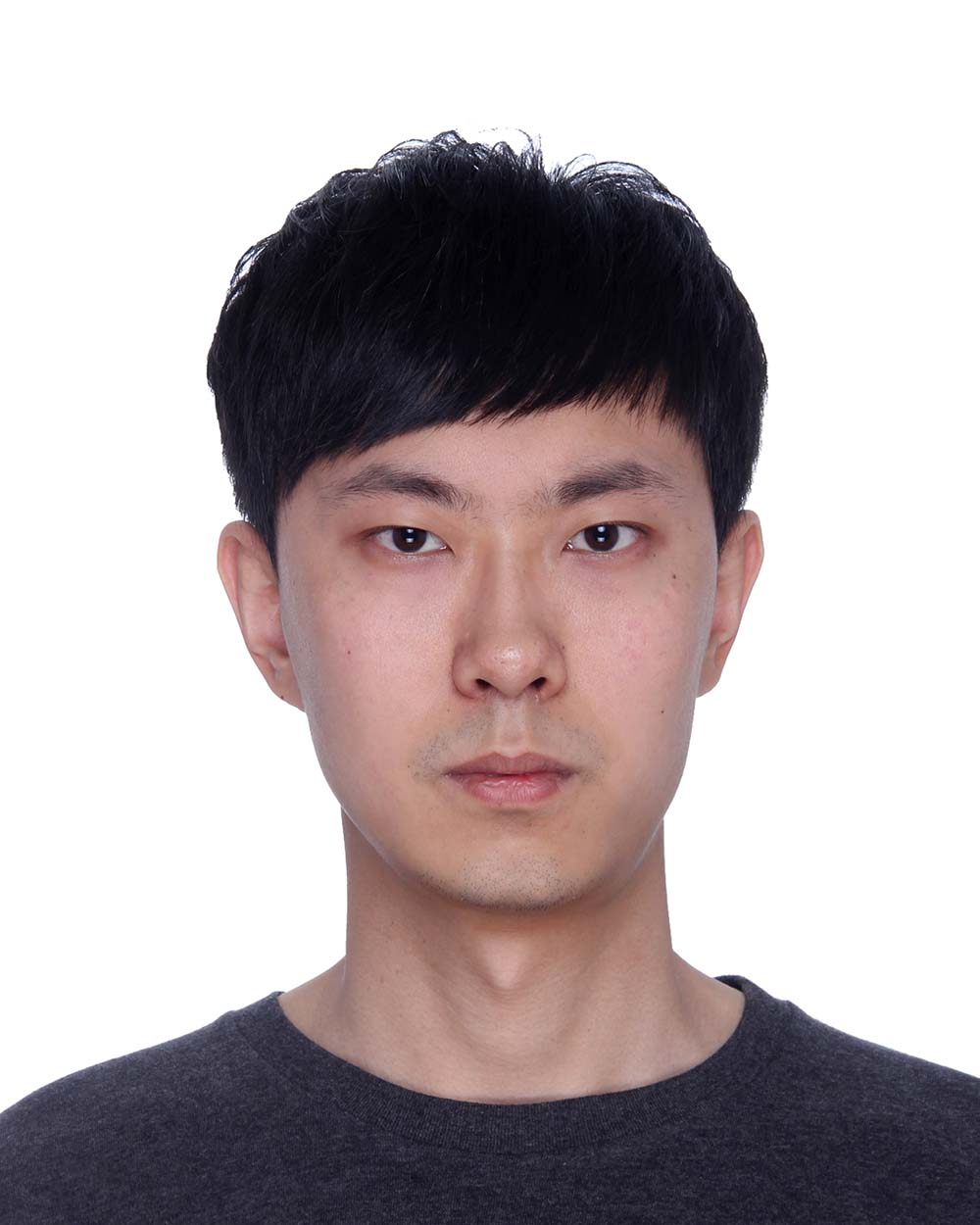}}]
{Xin Guo} received the B.S. degree in Mathematics and Applied Mathematics from Zhejiang University, Hangzhou, China, in 2015. He is now a Ph.D. candidate at the College of Computer Science and Technology, Zhejiang University, and also the State Key Laboratory of CAD \& CG. He is currently a research intern at Alibaba DAMO Academy. His research interests include computer vision and machine learning.
\end{IEEEbiography}

\begin{IEEEbiography}
[{\includegraphics[width=1in,height=1.25in]{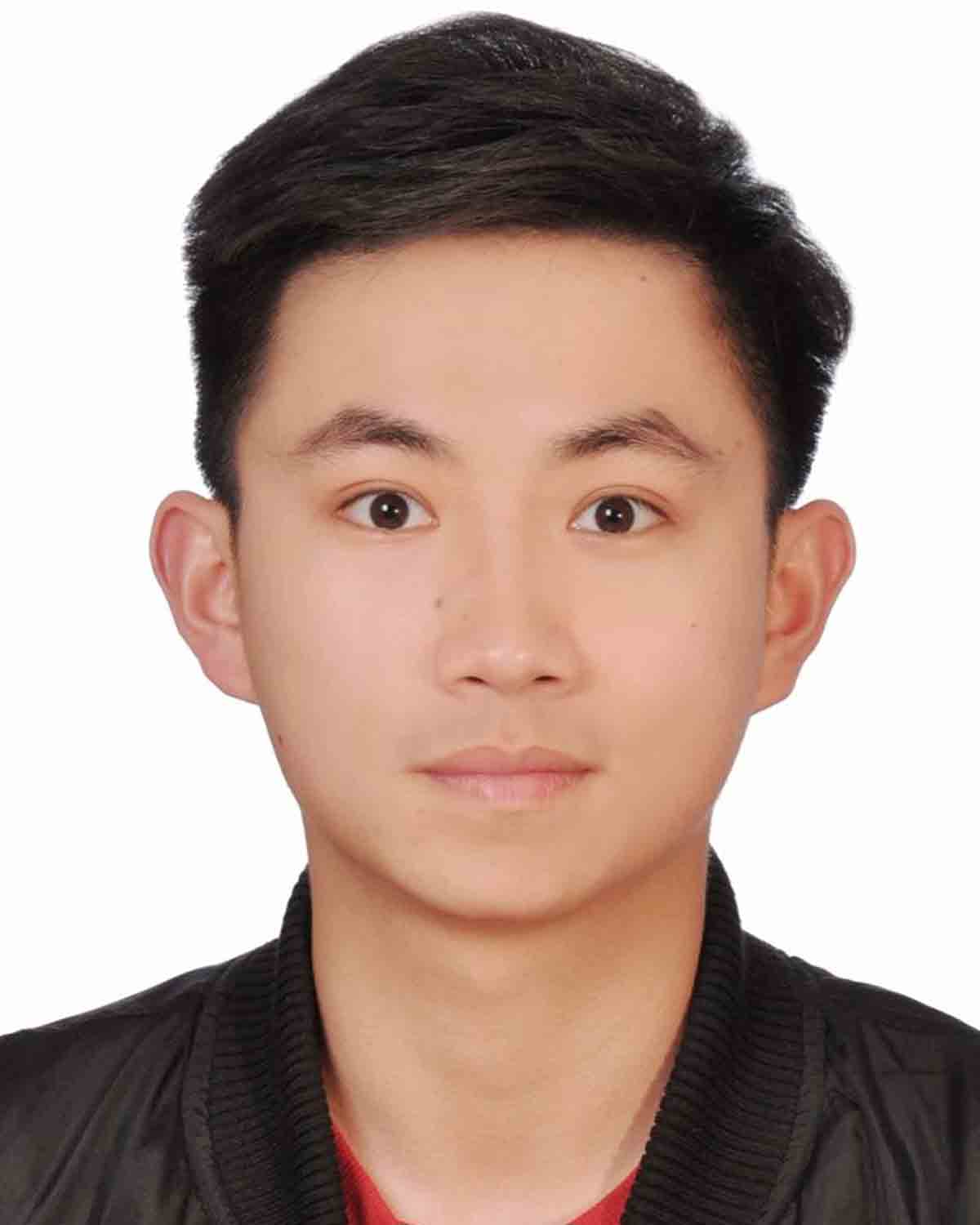}}]
{Zhengxu Yu} received the Ph.D. degree from Zhejiang University. He is currently an algorithm engineer with Alibaba DAMO Academy. His research interests include large-scale machine learning and computer vision.
\end{IEEEbiography}

\begin{IEEEbiography}
[{\includegraphics[width=1in,height=1.25in]{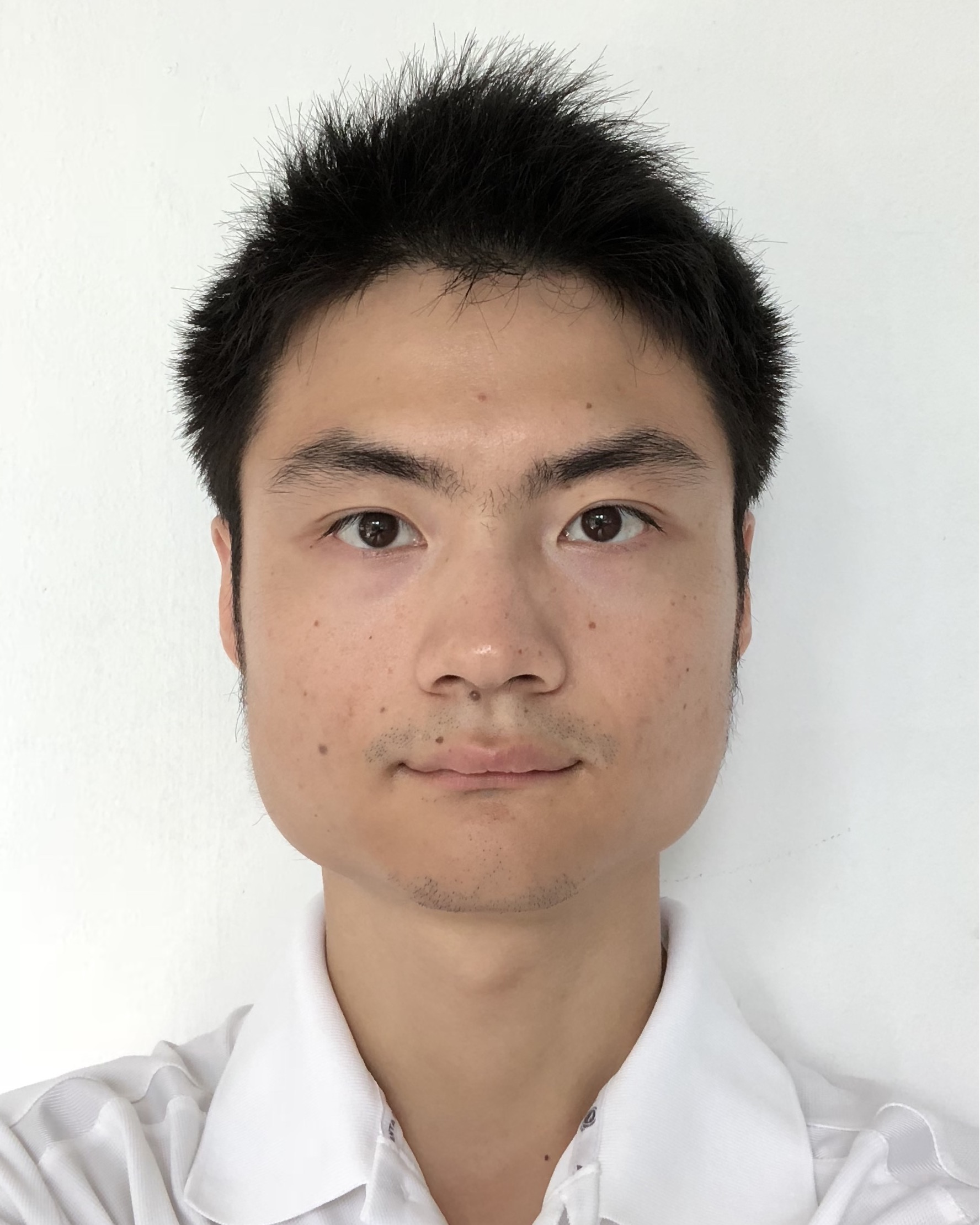}}]
{Chao Xiang} received the B.S. degree in Math and Applied Mathematics from Zhejiang University, China, in 2016. He is currently a fifth-year Ph.D. student in computer science at Zhejiang University. His research interests include machine learning, knowledge graph, and data mining.
\end{IEEEbiography}

\begin{IEEEbiography}
[{\includegraphics[width=1in,height=1.25in]{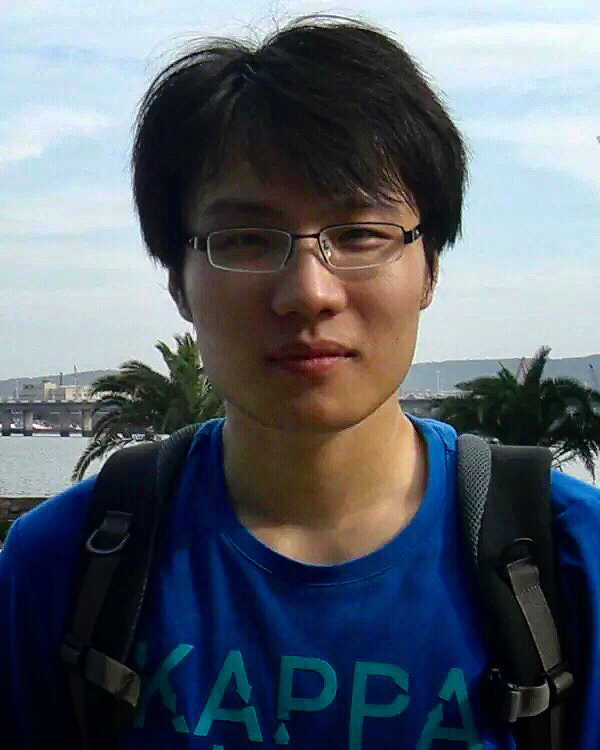}}]
{Zhongming Jin} is now a staff algorithm engineer at Alibaba DAMO Academy. Previously, he was a researcher at Baidu Research. He received his Ph.D. degree from Zhejiang University in Mar. 2015. His research interests include large-scale machine learning and computer vision.
\end{IEEEbiography}

\begin{IEEEbiography}
[{\includegraphics[width=1in,height=1.25in]{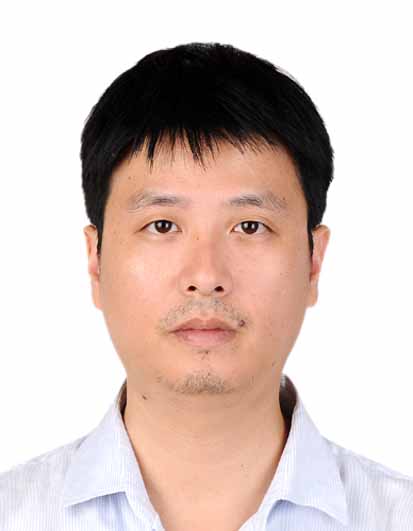}}]
{Jianqiang Huang} is a director of Alibaba DAMO Academy. He received the second prize in the National Science and Technology Progress Award in 2010. His research interests focus on visual intelligence in the city brain project of Alibaba.
\end{IEEEbiography}

\begin{IEEEbiography}
[{\includegraphics[width=1in,height=1.25in]{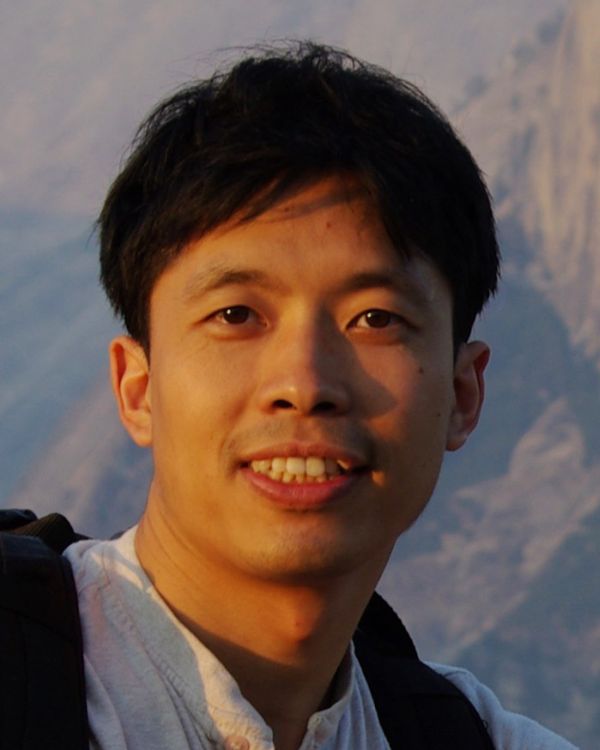}}]
{Deng Cai} is a professor in the State Key Lab of CAD\&CG, College of Computer Science at Zhejiang University, China. He received a Ph.D. degree in computer science from the University of Illinois at Urbana Champaign in 2009. His research interests include machine learning, data mining, and information retrieval.
\end{IEEEbiography}

\begin{IEEEbiography}
[{\includegraphics[width=1in,height=1.25in]{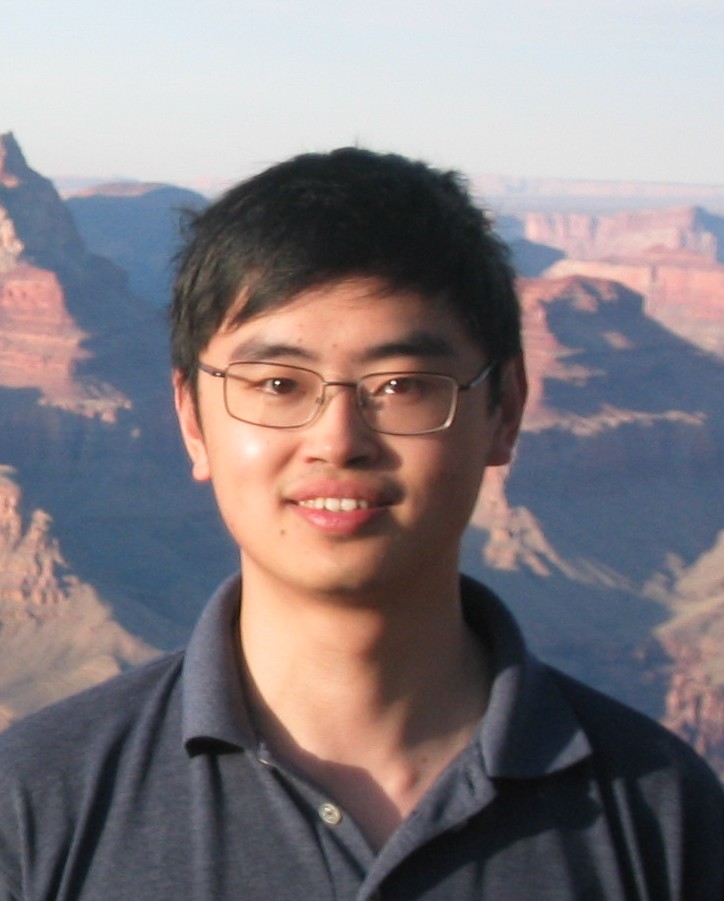}}]
{Xiaofei He} received a B.S. degree in Computer Science from Zhejiang University, China, in 2000 and a Ph.D. degree in Computer Science from the University of Chicago, in 2005. He is a Professor in the State Key Lab of CAD\&CG at Zhejiang University, China. Prior to joining Zhejiang University, he was a Research Scientist at Yahoo! Research Labs, Burbank, CA. His research interests include machine learning, information retrieval, and computer vision. He is a senior member of IEEE.
\end{IEEEbiography}

\begin{IEEEbiography}
[{\includegraphics[width=1in,height=1.25in]{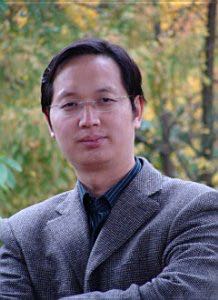}}]
{Xian-Sheng Hua} (F'16) received the B.S. and Ph.D. degrees in applied mathematics from Peking University, Beijing, in 1996 and 2001, respectively. In 2001, he joined Microsoft Research Asia as a Researcher and has been a Senior Researcher of Microsoft Research Redmond since 2013. He became a Researcher and the Senior Director of the Alibaba Group in 2015. He has authored or co-authored over 250 research papers and has filed over 90 patents. His research interests include multimedia search, advertising, understanding and mining, pattern recognition, and machine learning. He was honored as one of the recipients of MIT35. He served as a Program Co-Chair for the IEEE ICME 2013, the ACM Multimedia 2012, the IEEE ICME 2012, and on the Technical Directions Board of the IEEE Signal Processing Society. He is an ACM Distinguished Scientist and IEEE Fellow. 
\end{IEEEbiography}

\vfill

\end{document}